\documentstyle[aclap,ifthen,calc]{article}

\begin{document}


\def\diatop[#1|#2]{{\setbox1=\hbox{{#1{}}}\setbox2=\hbox{{#2{}}}%
                    \dimen0=\ifdim\wd1>\wd2\wd1\else\wd2\fi%
                    \dimen1=\ht2\advance\dimen1by-1ex%
                    \setbox1=\hbox to1\dimen0{\hss#1\hss}%
                    \rlap{\raise1\dimen1\box1}%
                    \hbox to1\dimen0{\hss#2\hss}}}%



\font\ipatwelverm=wsuipa12
\def\ipa{\ipatwelverm}

\def\inva{\edef\next{\the\font}\ipa\char'000\next}%
\def\scripta{\edef\next{\the\font}\ipa\char'001\next}%
\def\nialpha{\edef\next{\the\font}\ipa\char'002\next}%
\def\invscripta{\edef\next{\the\font}\ipa\char'003\next}%
\def\invv{\edef\next{\the\font}\ipa\char'004\next}%

\def\crossb{\edef\next{\the\font}\ipa\char'005\next}%
\def\barb{\edef\next{\the\font}\ipa\char'006\next}%
\def\slashb{\edef\next{\the\font}\ipa\char'007\next}%
\def\hookb{\edef\next{\the\font}\ipa\char'010\next}%
\def\nibeta{\edef\next{\the\font}\ipa\char'011\next}%

\def\slashc{\edef\next{\the\font}\ipa\char'012\next}%
\def\curlyc{\edef\next{\the\font}\ipa\char'013\next}%
\def\clickc{\edef\next{\the\font}\ipa\char'014\next}%

\def\crossd{\edef\next{\the\font}\ipa\char'015\next}%
\def\bard{\edef\next{\the\font}\ipa\char'016\next}%
\def\slashd{\edef\next{\the\font}\ipa\char'017\next}%
\def\hookd{\edef\next{\the\font}\ipa\char'020\next}%
\def\taild{\edef\next{\the\font}\ipa\char'021\next}%
\def\dz{\edef\next{\the\font}\ipa\char'022\next}%
\def\eth{\edef\next{\the\font}\ipa\char'023\next}%
\def\scd{\edef\next{\the\font}\ipa\char'024\next}%

\def\schwa{\edef\next{\the\font}\ipa\char'025\next}%
\def\er{\edef\next{\the\font}\ipa\char'026\next}%
\def\reve{\edef\next{\the\font}\ipa\char'027\next}%
\def\niepsilon{\edef\next{\the\font}\ipa\char'030\next}%
\def\revepsilon{\edef\next{\the\font}\ipa\char'031\next}%
\def\hookrevepsilon{\edef\next{\the\font}\ipa\char'032\next}%
\def\closedrevepsilon{\edef\next{\the\font}\ipa\char'033\next}%

\def\scriptg{\edef\next{\the\font}\ipa\char'034\next}%
\def\hookg{\edef\next{\the\font}\ipa\char'035\next}%
\def\scg{\edef\next{\the\font}\ipa\char'036\next}%
\def\nigamma{\edef\next{\the\font}\ipa\char'037\next}
\def\ipagamma{\edef\next{\the\font}\ipa\char'040\next}%
\def\babygamma{\edef\next{\the\font}\ipa\char'041\next}%

\def\hv{\edef\next{\the\font}\ipa\char'042\next}%
\def\crossh{\edef\next{\the\font}\ipa\char'043\next}%
\def\hookh{\edef\next{\the\font}\ipa\char'044\next}%
\def\hookheng{\edef\next{\the\font}\ipa\char'045\next}%
\def\invh{\edef\next{\the\font}\ipa\char'046\next}%

\def\bari{\edef\next{\the\font}\ipa\char'047\next}%
\def\dlbari{\edef\next{\the\font}\ipa\char'050\next}
\def\niiota{\edef\next{\the\font}\ipa\char'051\next}%
\def\sci{\edef\next{\the\font}\ipa\char'052\next}%
\def\barsci{\edef\next{\the\font}\ipa\char'053\next}

\def\invf{\edef\next{\the\font}\ipa\char'054\next}%

\def\tildel{\edef\next{\the\font}\ipa\char'055\next}%
\def\barl{\edef\next{\the\font}\ipa\char'056\next}%
\def\latfric{\edef\next{\the\font}\ipa\char'057\next}%
\def\taill{\edef\next{\the\font}\ipa\char'060\next}%
\def\lz{\edef\next{\the\font}\ipa\char'061\next}%
\def\nilambda{\edef\next{\the\font}\ipa\char'062\next}%
\def\crossnilambda{\edef\next{\the\font}\ipa\char'063\next}%

\def\labdentalnas{\edef\next{\the\font}\ipa\char'064\next}%
\def\invm{\edef\next{\the\font}\ipa\char'065\next}%
\def\legm{\edef\next{\the\font}\ipa\char'066\next}%

\def\nj{\edef\next{\the\font}\ipa\char'067\next}%
\def\eng{\edef\next{\the\font}\ipa\char'070\next}%
\def\tailn{\edef\next{\the\font}\ipa\char'071\next}%
\def\scn{\edef\next{\the\font}\ipa\char'072\next}%

\def\clickb{\edef\next{\the\font}\ipa\char'073\next}%
\def\baro{\edef\next{\the\font}\ipa\char'074\next}%
\def\openo{\edef\next{\the\font}\ipa\char'075\next}%
\def\niomega{\edef\next{\the\font}\ipa\char'076\next}%
\def\closedniomega{\edef\next{\the\font}\ipa\char'077\next}%
\def\oo{\edef\next{\the\font}\ipa\char'100\next}%

\def\barp{\edef\next{\the\font}\ipa\char'101\next}%
\def\thorn{\edef\next{\the\font}\ipa\char'102\next}%
\def\niphi{\edef\next{\the\font}\ipa\char'103\next}%

\def\flapr{\edef\next{\the\font}\ipa\char'104\next}%
\def\legr{\edef\next{\the\font}\ipa\char'105\next}%
\def\tailr{\edef\next{\the\font}\ipa\char'106\next}%
\def\invr{\edef\next{\the\font}\ipa\char'107\next}%
\def\tailinvr{\edef\next{\the\font}\ipa\char'110\next}%
\def\invlegr{\edef\next{\the\font}\ipa\char'111\next}%
\def\scr{\edef\next{\the\font}\ipa\char'112\next}%
\def\invscr{\edef\next{\the\font}\ipa\char'113\next}%

\def\tails{\edef\next{\the\font}\ipa\char'114\next}%
\def\esh{\edef\next{\the\font}\ipa\char'115\next}%
\def\curlyesh{\edef\next{\the\font}\ipa\char'116\next}%
\def\nisigma{\edef\next{\the\font}\ipa\char'117\next}%

\def\tailt{\edef\next{\the\font}\ipa\char'120\next}%
\def\tesh{\edef\next{\the\font}\ipa\char'121\next}%
\def\clickt{\edef\next{\the\font}\ipa\char'122\next}%
\def\nitheta{\edef\next{\the\font}\ipa\char'123\next}%

\def\baru{\edef\next{\the\font}\ipa\char'124\next}%
\def\slashu{\edef\next{\the\font}\ipa\char'125\next}%
\def\niupsilon{\edef\next{\the\font}\ipa\char'126\next}%
\def\scu{\edef\next{\the\font}\ipa\char'127\next}%
\def\barscu{\edef\next{\the\font}\ipa\char'130\next}%

\def\scriptv{\edef\next{\the\font}\ipa\char'131\next}%

\def\invw{\edef\next{\the\font}\ipa\char'132\next}%

\def\nichi{\edef\next{\the\font}\ipa\char'133\next}%

\def\invy{\edef\next{\the\font}\ipa\char'134\next}%
\def\scy{\edef\next{\the\font}\ipa\char'135\next}%

\def\curlyz{\edef\next{\the\font}\ipa\char'136\next}%
\def\tailz{\edef\next{\the\font}\ipa\char'137\next}%
\def\yogh{\edef\next{\the\font}\ipa\char'140\next}%
\def\curlyyogh{\edef\next{\the\font}\ipa\char'141\next}%

\def\glotstop{\edef\next{\the\font}\ipa\char'142\next}%
\def\revglotstop{\edef\next{\the\font}\ipa\char'143\next}%
\def\invglotstop{\edef\next{\the\font}\ipa\char'144\next}%
\def\ejective{\edef\next{\the\font}\ipa\char'145\next}%
\def\reveject{\edef\next{\the\font}\ipa\char'146\next}%


\def\dental#1{\oalign{#1\crcr
          \hidewidth{\ipa\char'147}\hidewidth}}

\def\upt{\edef\next{\the\font}\ipa\char'154\next}
\def\downt{\edef\next{\the\font}\ipa\char'155\next}%
\def\leftt{\edef\next{\the\font}\ipa\char'156\next}%
\def\rightt{\edef\next{\the\font}\ipa\char'157\next}%

\def\upp{\edef\next{\the\font}\ipa\char'164\next}
\def\downp{\edef\next{\the\font}\ipa\char'165\next}%
\def\leftp{\edef\next{\the\font}\ipa\char'166\next}%
\def\rightp{\edef\next{\the\font}\ipa\char'167\next}%

\def\stress{\edef\next{\the\font}\ipa\char'150\next}
\def\secstress{\edef\next{\the\font}\ipa\char'151\next}

\def\syllabic{\edef\next{\the\font}\ipa\char'152\next}

\def\corner{\edef\next{\the\font}\ipa\char'153\next}%

\def\halflength{\edef\next{\the\font}\ipa\char'160\next}
\def\length{\edef\next{\the\font}\ipa\char'161\next}

\def\underdots{\edef\next{\the\font}\ipa\char'162\next}%

\def\ain{\edef\next{\the\font}\ipa\char'163\next}

\def\overring{\edef\next{\the\font}\ipa\char'170\next}%
\def\underring{\edef\next{\the\font}\ipa\char'171\next}%

\def\open{\edef\next{\the\font}\ipa\char'172\next}%

\def\midtilde{\edef\next{\the\font}\ipa\char'173\next}%
\def\undertilde{\edef\next{\the\font}\ipa\char'174\next}%

\def\underwedge{\edef\next{\the\font}\ipa\char'175\next}%

\def\polishhook{\edef\next{\the\font}\ipa\char'176\next}%

\def\underarch#1{\oalign{#1\crcr
          \hidewidth{\ipa\char'177}\hidewidth}}


\font\ipatenrm=wsuipa10
\def\ipa{\ipatenrm}


\newcommand{\Syl}{$\sigma$}             
\newcommand{\Mor}{$\mu$}                
\newcommand{\Sylm}{$\sigma_{\mu}$}      
\newcommand{\Sylmm}{$\sigma_{\mu\mu}$}  
\newcommand{\Sylx}{$\sigma_{x}$} 
\newcommand{\kernel}{B:$\Phi$}          
\newcommand{\residue}{B/$\Phi$}         
\newcommand{\ppc}{O:$\Phi$}             
\newcommand{\npc}{O/$\Phi$}             

\newcommand{\composition}{$\circ$}      
\newcommand{\conc}{$^{\frown}$}         
\newcommand{\estr}{$\varepsilon$}   
\newcommand{\func}[2]
   {\mbox{{\sc #1(}#2{\sc )}}}

\newcommand{\lab}{$\langle$} 
\newcommand{\rab}{$\rangle$} 


\newcounter{boxwidth}
\newcounter{boxhight}
\newcounter{mttlmboxwidth}
\newcounter{mttlmboxhight}
\newcounter{autosegboxwidth}
\newcounter{defaulthight}
\newcounter{strlength}
\newcounter{notiers}
\newcounter{picwidth}
\newcounter{pichight}
\newcounter{lift}
\newcounter{linelen}
\newcounter{xdirection}
\newcounter{ydirection}
\newcounter{curx}
\newcounter{cury}
\newcounter{pictopmargin}

\newlength{\tapenamewidth}
\newlength{\templen}

\setcounter{mttlmboxwidth}{12}%
\setcounter{mttlmboxhight}{12}
\setcounter{autosegboxwidth}{10}%
\setcounter{pictopmargin}{2}

\newcounter{fsmboxwidth}  \setcounter{fsmboxwidth}{30}
\newcounter{fsmcolumns}  \setcounter{fsmcolumns}{3}
\newcounter{fsmrows}
\newcounter{fsmcolumnsx}
\newcounter{fsmrowsx}
\newcounter{nostates}
\newcounter{curstate}
\newcounter{fsmradius}

\newcommand{\fsm}[2]
   {
    \setcounter{boxwidth}{\value{fsmboxwidth}}%
    \setcounter{boxhight}{\value{fsmboxwidth}}%
    \setcounter{nostates}{0}%
    \countstates#1|| END%
    \setcounter{fsmrows}{\value{nostates}/\value{fsmcolumns}}%
    \setcounter{temp}{\value{fsmrows}*\value{fsmcolumns}}%
    \ifthenelse{\value{nostates} = \value{temp}}%
               {}%
               {\stepcounter{fsmrows}}%
    \setcounter{fsmcolumnsx}{\value{fsmcolumns}-1}%
    \setcounter{fsmrowsx}{\value{fsmrows}-1}%
    \setcounter{fsmradius}{\value{boxwidth}-\value{boxwidth}/10}%
    \setcounter{picwidth}{\value{boxwidth}*\value{fsmcolumns}+%
                          \value{boxwidth}*\value{fsmcolumnsx}+%
                          \value{boxwidth}*2}%
    \setcounter{pichight}{\value{boxhight}*\value{fsmrows}+%
                          \value{boxhight}*\value{fsmrowsx}+%
                          \value{boxhight}*2}%
    \setcounter{lift}{\value{pichight}/-2}%
    \rule[\the\value{lift}pt]{3 pt}{\the\value{pichight}pt}%
    \begin{picture}(\the\value{picwidth},0)(0,-\the\value{lift})%
       \setcounter{curstate}{0}%
       \setcounter{curx}{\value{boxhight}}%
       \setcounter{cury}{\value{boxhight}*\value{fsmrowsx}*2+\value{boxhight}}%
       \drawstates#1|| END%
    \end{picture}
   }

\def\drawstates#1,#2,#3|#4 END
   {
    \put(\the\value{curx},\the\value{cury})%
         {\makebox(\the\value{boxwidth},\the\value{boxhight}){#1}}%
    \setcounter{curx}{\value{curx}+\value{boxwidth}/2}%
    \setcounter{cury}{\value{cury}+\value{boxhight}/2}%
    \put(\the\value{curx},\the\value{cury})%
         {\circle{\the\value{boxwidth}}}%
    \ifthenelse{\equal{#3}{y}}%
       {\put(\the\value{curx},\the\value{cury})%
            {\circle{\the\value{fsmradius}}}}%
       {}%
    \setcounter{curx}{\value{curx}-\value{boxwidth}/2}%
    \setcounter{cury}{\value{cury}-\value{boxhight}/2}%
    \stepcounter{curstate}%
    \ifthenelse{\value{curstate} = \value{fsmcolumns}}
               {\setcounter{curx}{\value{boxhight}}%
                \setcounter{cury}{\value{cury}-\value{boxhight}*2}%
                \setcounter{curstate}{0}}%
               {\setcounter{curx}{\value{curx}+\value{boxwidth}*2}}%
    \ifthenelse{\equal{#4}{|}}{}%
               {\drawstates#4 END}%
   }

\def\countstates#1|#2 END
   {\stepcounter{nostates}%
    \ifthenelse{\equal{#2}{|}}{}%
               {\countstates#2 END}%
   }


\newcommand{\mttlmsetwidth}[1]%
   {\setcounter{mttlmboxwidth}{#1}}

\newcommand{\mttlmsethight}[1]%
   {\setcounter{mttlmboxhight}{#1}}

\newcommand{\mttlm}[3]
   {\immediate\write16{MTTLM = #1}%
    \setcounter{boxwidth}{\value{mttlmboxwidth}}%
    \setcounter{boxhight}{\value{mttlmboxhight}}%
    \setcounter{strlength}{0}%
    \setcounter{notiers}{0}%
    \setlength{\tapenamewidth}{0pt}%
    \counttiers#3|| END%
    \countsurfacewidth#1 END%
    \findtapenamewidth#3|| END%
    \setcounter{picwidth}{\value{boxwidth}*\value{strlength}}%
    \setcounter{pichight}{\value{boxhight}*\value{notiers}+%
                          2*\value{boxhight}+\value{pictopmargin}}%
    \setcounter{lift}{\value{pichight}/-2}%
    \rule[\the\value{lift}pt]{0 pt}{\the\value{pichight}pt}%
    \begin{picture}(\the\value{picwidth},0)(0,-\the\value{lift})%
       \let\boxtype=\makebox%
       \setcounter{curx}{0}%
       \setcounter{cury}{0}%
       \displayonetape#1 END%
       \setcounter{curx}{0}%
       \setcounter{cury}{\value{boxhight}}%
       \displaystrings#2|| END%
       \setcounter{curx}{0}%
       \setcounter{cury}{2*\value{boxhight}}%
       \displaymanytapes#3|| END%
    \end{picture}%
    \hspace{\the\tapenamewidth}%
   }

\def\countsurfacewidth#1:#2 END%
   {\countwidth#1| END%
    \maxtapenamewidth#2 END%
   }

\def\displayonetape#1:#2 END%
   {
    \put(\the\value{curx},\the\value{cury})%
         {\framebox(\the\value{picwidth},\the\value{boxhight}){}}%
    \partition#1| END%
    \put(\the\value{curx},\the\value{cury})%
         {\makebox(\the\value{boxwidth},\the\value{boxhight})[l]{\ {\em #2}}}%
    \setcounter{curx}{0}%
    \displaystrings#1|| END%
   }

\def\displaymanytapes#1|#2 END%
   {\displayonetape#1 END%
    \addtocounter{cury}{-\value{boxhight}}%
    \ifthenelse{\equal{#2}{|}}{}%
               {\displaymanytapes#2 END}%
   }

\def\partition#1#2 END%
   {\addtocounter{curx}{\value{boxwidth}}%
    \ifthenelse{\equal{#1}{-}}{}%
                      {\put(\value{curx},\value{cury})%
                         {\dashbox{0.5}(0,\value{boxhight}){}}}
    \ifthenelse{\equal{#2}{|}}{}%
               {\partition#2 END}%
   }

\def\findtapenamewidth#1:#2|#3 END
   {\maxtapenamewidth#2 END%
    \ifthenelse{\equal{#3}{|}}{}%
               {\findtapenamewidth#3 END}%
   }

\def\maxtapenamewidth#1 END%
   {\settowidth{\templen}{{\em #1}}%
    \ifthenelse{\templen > \tapenamewidth}%
               {\settowidth{\tapenamewidth}{\ {\em #1}}}%
               {}%
   }


\newcommand{\autosegsetwidth}[1]%
   {\setcounter{autosegboxwidth}{#1}}

\newcommand{\autoseg}[2]%
   {\immediate\write16{Autseg Tiers #1}%
    \setcounter{boxwidth}{\value{autosegboxwidth}}%
    \setcounter{boxhight}{\value{autosegboxwidth}}%
    \setcounter{strlength}{0}%
    \setcounter{notiers}{0}%
    \counttiers#1|| END%
    \countwidth#1 END%
    \setcounter{picwidth}{\value{boxwidth}*\value{strlength}}%
    \setcounter{pichight}{2*\value{boxwidth}*\value{notiers}-%
                          \value{boxwidth}+\value{pictopmargin}}%
    \setcounter{lift}{\value{pichight}/-2}%
    \rule[\the\value{lift}pt]{0 pt}{\the\value{pichight}pt}%
    \begin{picture}(\the\value{picwidth},0)(0,-\the\value{lift})%
       \let\boxtype=\makebox%
       \setcounter{curx}{0}%
       \setcounter{cury}{0}%
       \displaystrings#1|| END%
       \setcounter{curx}{\value{boxwidth}/2}%
       \linkstrings#2,||| END%
    \end{picture}%
   }

\def\counttiers#1|#2 END
   {\stepcounter{notiers}%
    \ifthenelse{\equal{#2}{|}}{}%
               {\counttiers#2 END}%
   }

\def\countwidth#1|#2 END%
   {\countlength#1| END%
   }

\def\countlength#1#2 END
   {\stepcounter{strlength}%
    \ifthenelse{\equal{#2}{|}}{}%
               {\countlength#2 END}%
   }

\def\displaystrings#1|#2 END
   {\setmorpheme#1| END%
    \setcounter{curx}{0}%
    \addtocounter{cury}{2*\value{boxhight}}%
    \ifthenelse{\equal{#2}{|}}{}%
               {\displaystrings#2 END}%
   }

\def\setmorpheme#1#2 END%
   {\ifthenelse{\equal{#1}{-}}{}%
               {\put(\the\value{curx},\the\value{cury})%
                {\boxtype(\the\value{boxwidth},\the\value{boxhight}){#1}}}%
    \addtocounter{curx}{\value{boxwidth}}%
    \ifthenelse{\equal{#2}{|}}{}%
               {\setmorpheme#2 END}%
   }

\def\linkstrings#1#2#3,#4 END
   {
    \ifthenelse{\equal{#1}{-}}{}%
       {
        \ifthenelse{#1 < #2}%
               {\setcounter{cury}{(2*#1-1)*\value{boxwidth}}%
                \setcounter{ydirection}{1}%
                \setcounter{linelen}{(2*(#2-#1)-1)*\value{boxwidth}}}%
               {}%
        \ifthenelse{#1 > #2}%
               {\setcounter{cury}{2*(#1-1)*\value{boxwidth}}%
                \setcounter{ydirection}{-1}%
                \setcounter{linelen}{(2*(#1-#2)-1)*\value{boxwidth}}}%
               {}%
        \setcounter{xdirection}{0}%
        \ifthenelse{\equal{#3}{r}}{\setcounter{xdirection}{1}}{}%
        \ifthenelse{\equal{#3}{rr}}{\setcounter{xdirection}{2}%
                                   \addtocounter{linelen}{\value{boxwidth}}}{}%
        \ifthenelse{\equal{#3}{rrr}}{\setcounter{xdirection}{3}%
                                   \addtocounter{linelen}{2*\value{boxwidth}}}%
                                   {}%
        \ifthenelse{\equal{#3}{l}}{\setcounter{xdirection}{-1}}{}%
        \ifthenelse{\equal{#3}{ll}}{\setcounter{xdirection}{-2}%
                                   \addtocounter{linelen}{\value{boxwidth}}}{}%
        \ifthenelse{\equal{#3}{lll}}{\setcounter{xdirection}{-3}%
                                   \addtocounter{linelen}{2*\value{boxwidth}}}%
                                   {}%
        \put(\value{curx},\value{cury})%
            {\line(\value{xdirection},\value{ydirection})%
            {\value{linelen}}}}%
    \addtocounter{curx}{\value{boxwidth}}%
    \ifthenelse{\equal{#4}{|||}}{}%
               {\linkstrings#4 END}%
   }

\newcounter{moraboxwidth}
\setcounter{moraboxwidth}{10}

\newcommand{\morasetwidth}[1]%
   {\setcounter{moraboxwidth}{#1}}

\newcommand{\moratree}[2]%
   {
    \setcounter{picwidth}{2*\value{moraboxwidth}}%
    \setcounter{pichight}{4*\value{moraboxwidth}+\value{moraboxwidth}/2}%
    \setcounter{lift}{\value{pichight}/-2}%
    \rule[\the\value{lift}pt]{0 pt}{\the\value{pichight}pt}%
    \begin{picture}(\the\value{picwidth},0)(0,-\the\value{lift})%
       \drawbasic{#1}{#2}%
    \end{picture}%
   }

\newcommand{\mmoratree}[3]%
   {
    \setcounter{picwidth}{3*\value{moraboxwidth}}%
    \setcounter{pichight}{4*\value{moraboxwidth}+\value{moraboxwidth}/2}%
    \setcounter{lift}{\value{pichight}/-2}%
    \rule[\the\value{lift}pt]{0 pt}{\the\value{pichight}pt}%
    \begin{picture}(\the\value{picwidth},0)(0,-\the\value{lift})%
       \drawbasic{#1}{#2}%
       \setcounter{curx}{2*\value{moraboxwidth}}%
       \put(\value{curx},0)%
          {\makebox(\value{moraboxwidth},\value{moraboxwidth})[b]{#3}}%
       \setcounter{cury}{2*\value{moraboxwidth}}%
       \put(\value{curx},\value{cury})%
          {\makebox(\value{moraboxwidth},\value{moraboxwidth}){\Mor}}%
       \setcounter{curx}{2*\value{moraboxwidth}+\value{moraboxwidth}/2}%
       \put(\value{curx},\value{cury}){\line(0,-1){\value{moraboxwidth}}}%
       \setcounter{curx}{\value{moraboxwidth}+\value{moraboxwidth}/2}%
       \setcounter{cury}{3*\value{moraboxwidth}+\value{moraboxwidth}/2}%
       \put(\value{curx},\value{cury}){\line(2,-1){\value{moraboxwidth}}}%
    \end{picture}%
   }

\newcommand{\xmoratree}[1]%
   {
    \setcounter{picwidth}{\value{moraboxwidth}}%
    \setcounter{pichight}{4*\value{moraboxwidth}+\value{moraboxwidth}/2}%
    \setcounter{lift}{\value{pichight}/-2}%
    \rule[\the\value{lift}pt]{0 pt}{\the\value{pichight}pt}%
    \begin{picture}(\the\value{picwidth},0)(0,-\the\value{lift})%
       \put(0,0){\makebox(\value{moraboxwidth},\value{moraboxwidth})[b]{#1}}%
       \setcounter{cury}{3*\value{moraboxwidth}+\value{moraboxwidth}/2}%
       \put(0,\value{cury})%
       {\makebox(\value{moraboxwidth},\value{moraboxwidth}){\Sylx}}%
       \setcounter{curx}{\value{moraboxwidth}/2}%
       \setcounter{linelen}{2*\value{moraboxwidth}+\value{moraboxwidth}/2}%
       \put(\value{curx},\value{cury}){\line(0,-1){\value{linelen}}}%
    \end{picture}%
   }

\newcommand{\gmoratree}[5]%
   {\mbox{\mmoratree{#1}{#2}{#3}%
          \hspace{-\value{moraboxwidth}pt}%
          \mmoratree{{}}{#4}{#5}}%
   \immediate\write16{(#1,#2,#3,#4,#5)}%
   }

\newcommand{\drawbasic}[2]%
   {
    \put(0,0){\makebox(\value{moraboxwidth},\value{moraboxwidth})[b]{#1}}%
    \put(\value{moraboxwidth},0)%
       {\makebox(\value{moraboxwidth},\value{moraboxwidth})[b]{#2}}%
    \setcounter{cury}{2*\value{moraboxwidth}}%
    \put(\value{moraboxwidth},\value{cury})%
       {\makebox(\value{moraboxwidth},\value{moraboxwidth}){\Mor}}%
    \setcounter{cury}{3*\value{moraboxwidth}+\value{moraboxwidth}/2}%
    \put(\value{moraboxwidth},\value{cury})%
       {\makebox(\value{moraboxwidth},\value{moraboxwidth}){\Syl}}%
    \setcounter{curx}{\value{moraboxwidth}+\value{moraboxwidth}/2}%
    \setcounter{cury}{3*\value{moraboxwidth}+\value{moraboxwidth}/2}%
    \put(\value{curx},\value{cury}){\line(-2,-5){\value{moraboxwidth}}}%
    \setcounter{linelen}{\value{moraboxwidth}/2}%
    \put(\value{curx},\value{cury}){\line(0,-1){\value{linelen}}}%
    \setcounter{cury}{2*\value{moraboxwidth}}%
    \put(\value{curx},\value{cury}){\line(0,-1){\value{moraboxwidth}}}%
   }

\newcounter{tapeboxhight}
\setcounter{tapeboxhight}{15}
\newcounter{delta}
\newcounter{fstwidth}
\newcounter{temp}

\newcommand{\tapehight}[1]%
   {\setcounter{tapeboxhight}{#1}}

\newcommand{\cascadetransducers}%
   {
    \setcounter{picwidth}{10*\value{tapeboxhight}}%
    \setcounter{pichight}{9*\value{tapeboxhight}}%
    \setcounter{lift}{\value{pichight}/-2}%
    \rule[\the\value{lift}pt]{0 pt}{\the\value{pichight}pt}%
    \begin{picture}(\the\value{picwidth},0)(0,-\the\value{lift})%
       \put(0,0){\framebox(\value{picwidth},\value{tapeboxhight})
             {Surface String}}%
       \setcounter{cury}{4*\value{tapeboxhight}}
       \setcounter{temp}{8*\value{tapeboxhight}}
       \put(\value{tapeboxhight},\value{cury})
             {\framebox(\value{temp},\value{tapeboxhight})
             {Intermediate String}}%
       \setcounter{cury}{8*\value{tapeboxhight}}
       \put(0,\value{cury}){\framebox(\value{picwidth},\value{tapeboxhight})
             {Lexical String}}%
       \setcounter{curx}{\value{picwidth}/2}
       \setcounter{cury}{\value{tapeboxhight}}
       \setcounter{delta}{2*\value{tapeboxhight}}
       \multiput(\value{curx},\value{cury})(0,\value{delta}){4}
          {\line(0,1){\value{tapeboxhight}}}
       \setcounter{cury}{2*\value{tapeboxhight}+\value{tapeboxhight}/2}
       \setcounter{delta}{4*\value{tapeboxhight}}
       \setcounter{fstwidth}{2*\value{tapeboxhight}}
       \multiput(\value{curx},\value{cury})(0,\value{delta}){2}
          {\oval(\value{fstwidth},\value{tapeboxhight})}
       \put(\value{curx},\value{cury}){\makebox(0,0){\em FST$_n$}}%
       \setcounter{cury}{6*\value{tapeboxhight}+\value{tapeboxhight}/2}
       \put(\value{curx},\value{cury}){\makebox(0,0){\em FST$_1$}}%
    \end{picture}%
    ~$\Longrightarrow$~%
    \begin{picture}(\the\value{picwidth},0)(0,-\the\value{lift})%
       \put(0,0){\framebox(\value{picwidth},\value{tapeboxhight})
             {Surface String}}%
       \setcounter{cury}{8*\value{tapeboxhight}}
       \put(0,\value{cury}){\framebox(\value{picwidth},\value{tapeboxhight})
             {Lexical String}}%
       \setcounter{curx}{\value{picwidth}/2}
       \setcounter{cury}{\value{tapeboxhight}}
       \setcounter{temp}{3*\value{tapeboxhight}}
       \setcounter{delta}{4*\value{tapeboxhight}}
       \multiput(\value{curx},\value{cury})(0,\value{delta}){2}
          {\line(0,1){\value{temp}}}
       \setcounter{cury}{4*\value{tapeboxhight}+\value{tapeboxhight}/2}
       \setcounter{fstwidth}{8*\value{tapeboxhight}}
       \put(\value{curx},\value{cury})
              {\oval(\value{fstwidth},\value{tapeboxhight})}
       \put(\value{curx},\value{cury})
              {\makebox(0,0){$FST_1 \circ FST_2 \circ \cdots \circ FST_n$}}%
    \end{picture}%
   }

\newcommand{\paralleltransducers}%
   {\paralleltransducersone%
    ~$\Longrightarrow$~%
    \paralleltransducerstwo%
   }

\newcommand{\paralleltransducersone}%
   {
    \setcounter{picwidth}{10*\value{tapeboxhight}}%
    \setcounter{pichight}{9*\value{tapeboxhight}}%
    \setcounter{lift}{\value{pichight}/-2}%
    \rule[\the\value{lift}pt]{0 pt}{\the\value{pichight}pt}%
    \begin{picture}(\the\value{picwidth},0)(0,-\the\value{lift})%
       \put(0,0){\framebox(\value{picwidth},\value{tapeboxhight})
             {Surface String}}%
       \setcounter{cury}{8*\value{tapeboxhight}}
       \put(0,\value{cury}){\framebox(\value{picwidth},\value{tapeboxhight})
             {Lexical String}}%
       \setcounter{curx}{\value{picwidth}/2}
       \setcounter{cury}{\value{tapeboxhight}}
       \setcounter{delta}{6*\value{tapeboxhight}}
       \multiput(\value{curx},\value{cury})(0,\value{delta}){2}
          {\line(0,1){\value{tapeboxhight}}}
       \setcounter{curx}{\value{tapeboxhight}}
       \setcounter{cury}{2*\value{tapeboxhight}}
       \setcounter{delta}{5*\value{tapeboxhight}}
       \setcounter{temp}{8*\value{tapeboxhight}}
       \multiput(\value{curx},\value{cury})(0,\value{delta}){2}
          {\line(1,0){\value{temp}}}
       \setcounter{delta}{3*\value{tapeboxhight}}
       \setcounter{temp}{2*\value{tapeboxhight}}
       \multiput(\value{curx},\value{cury})(\value{delta},0){2}
          {\line(0,1){\value{temp}}}
       \setcounter{cury}{5*\value{tapeboxhight}}
       \multiput(\value{curx},\value{cury})(\value{delta},0){2}
          {\line(0,1){\value{temp}}}
       \setcounter{cury}{4*\value{tapeboxhight}+\value{tapeboxhight}/2}
       \setcounter{fstwidth}{2*\value{tapeboxhight}}
       \multiput(\value{curx},\value{cury})(\value{delta},0){2}
          {\oval(\value{fstwidth},\value{tapeboxhight})}
       \put(\value{curx},\value{cury}){\makebox(0,0){\em FST$_1$}}%
       \setcounter{curx}{4*\value{tapeboxhight}}
       \put(\value{curx},\value{cury}){\makebox(0,0){\em FST$_2$}}%
       \setcounter{curx}{9*\value{tapeboxhight}}
       \setcounter{cury}{2*\value{tapeboxhight}}
       \setcounter{delta}{3*\value{tapeboxhight}}
       \multiput(\value{curx},\value{cury})(0,\value{delta}){2}
          {\line(0,1){\value{temp}}}
       \setcounter{cury}{4*\value{tapeboxhight}+\value{tapeboxhight}/2}
       \put(\value{curx},\value{cury})
          {\oval(\value{fstwidth},\value{tapeboxhight})}
       \put(\value{curx},\value{cury}){\makebox(0,0){\em FST$_n$}}%
       \setcounter{curx}{6*\value{tapeboxhight}+\value{tapeboxhight}/2}
       \put(\value{curx},\value{cury}){\makebox(0,0){$\cdots$}}%
    \end{picture}%
   }
\newcommand{\paralleltransducerstwo}%
   {\begin{picture}(\the\value{picwidth},0)(0,-\the\value{lift})%
       \put(0,0){\framebox(\value{picwidth},\value{tapeboxhight})
             {Surface String}}%
       \setcounter{cury}{8*\value{tapeboxhight}}
       \put(0,\value{cury}){\framebox(\value{picwidth},\value{tapeboxhight})
             {Lexical String}}%
       \setcounter{curx}{\value{picwidth}/2}
       \setcounter{cury}{\value{tapeboxhight}}
       \setcounter{temp}{3*\value{tapeboxhight}}
       \setcounter{delta}{4*\value{tapeboxhight}}
       \multiput(\value{curx},\value{cury})(0,\value{delta}){2}
          {\line(0,1){\value{temp}}}
       \setcounter{cury}{4*\value{tapeboxhight}+\value{tapeboxhight}/2}
       \setcounter{fstwidth}{8*\value{tapeboxhight}}
       \put(\value{curx},\value{cury})
              {\oval(\value{fstwidth},\value{tapeboxhight})}
       \put(\value{curx},\value{cury})
              {\makebox(0,0){$FST_1 \cap FST_2 \cap \cdots \cap FST_n$}}%
    \end{picture}%
   }

\newcommand{\uniontransducers}%
   {
    \setcounter{picwidth}{10*\value{tapeboxhight}}%
    \setcounter{pichight}{9*\value{tapeboxhight}}%
    \setcounter{lift}{\value{pichight}/-2}%
    \rule[\the\value{lift}pt]{0 pt}{\the\value{pichight}pt}%
    \begin{picture}(\the\value{picwidth},0)(0,-\the\value{lift})%
       \put(0,0){\framebox(\value{picwidth},\value{tapeboxhight})
             {Surface String}}%
       \setcounter{curx}{2*\value{tapeboxhight}+\value{tapeboxhight}/2}
       \setcounter{delta}{2*\value{tapeboxhight}+\value{tapeboxhight}/2}
       \multiput(\value{curx},0)(\value{delta},0){3}
          {\dashbox{.75}(0,\value{tapeboxhight}){}}
       \setcounter{cury}{8*\value{tapeboxhight}}
       \put(0,\value{cury}){\framebox(\value{picwidth},\value{tapeboxhight})
             {Lexical String}}%
       \setcounter{cury}{8*\value{tapeboxhight}}
       \multiput(\value{curx},\value{cury})(\value{delta},0){3}
          {\dashbox{.75}(0,\value{tapeboxhight}){}}
       \setcounter{curx}{\value{tapeboxhight}}
       \setcounter{cury}{\value{tapeboxhight}}
       \setcounter{delta}{3*\value{tapeboxhight}}
       \multiput(\value{curx},\value{cury})(\value{delta},0){2}
          {\line(0,1){\value{delta}}}
       \setcounter{cury}{5*\value{tapeboxhight}}
       \multiput(\value{curx},\value{cury})(\value{delta},0){2}
          {\line(0,1){\value{delta}}}
       \setcounter{cury}{4*\value{tapeboxhight}+\value{tapeboxhight}/2}
       \setcounter{fstwidth}{2*\value{tapeboxhight}}
       \multiput(\value{curx},\value{cury})(\value{delta},0){2}
          {\oval(\value{fstwidth},\value{tapeboxhight})}
       \put(\value{curx},\value{cury}){\makebox(0,0){\em FST$_1$}}%
       \setcounter{curx}{4*\value{tapeboxhight}}
       \put(\value{curx},\value{cury}){\makebox(0,0){\em FST$_2$}}%
       \setcounter{curx}{9*\value{tapeboxhight}}
       \setcounter{cury}{\value{tapeboxhight}}
       \setcounter{delta}{4*\value{tapeboxhight}}
       \multiput(\value{curx},\value{cury})(0,\value{delta}){2}
          {\line(0,1){\value{temp}}}
       \setcounter{cury}{4*\value{tapeboxhight}+\value{tapeboxhight}/2}
       \put(\value{curx},\value{cury})
          {\oval(\value{fstwidth},\value{tapeboxhight})}
       \put(\value{curx},\value{cury}){\makebox(0,0){\em FST$_n$}}%
       \setcounter{curx}{6*\value{tapeboxhight}+\value{tapeboxhight}/2}
       \put(\value{curx},\value{cury}){\makebox(0,0){$\cdots$}}%
    \end{picture}%
    ~$\Longrightarrow$~%
    \begin{picture}(\the\value{picwidth},0)(0,-\the\value{lift})%
       \put(0,0){\framebox(\value{picwidth},\value{tapeboxhight})
             {Surface String}}%
       \setcounter{curx}{2*\value{tapeboxhight}+\value{tapeboxhight}/2}
       \setcounter{delta}{2*\value{tapeboxhight}+\value{tapeboxhight}/2}
       \multiput(\value{curx},0)(\value{delta},0){3}
          {\dashbox{.75}(0,\value{tapeboxhight}){}}
       \setcounter{cury}{8*\value{tapeboxhight}}
       \put(0,\value{cury}){\framebox(\value{picwidth},\value{tapeboxhight})
             {Lexical String}}%
       \setcounter{cury}{8*\value{tapeboxhight}}
       \multiput(\value{curx},\value{cury})(\value{delta},0){3}
          {\dashbox{.75}(0,\value{tapeboxhight}){}}
       \setcounter{curx}{\value{picwidth}/2}
       \setcounter{cury}{4*\value{tapeboxhight}+\value{tapeboxhight}/2}
       \setcounter{fstwidth}{8*\value{tapeboxhight}}
       \put(\value{curx},\value{cury})
              {\oval(\value{fstwidth},\value{tapeboxhight})}
       \put(\value{curx},\value{cury})
              {\makebox(0,0){$FST_1 \cup FST_2 \cup \cdots \cup FST_n$}}%
       \setcounter{cury}{5*\value{tapeboxhight}}
       \setcounter{temp}{4*\value{tapeboxhight}}
       \put(\value{curx},\value{cury}){\line(-4,3){\value{temp}}}
       \put(\value{curx},\value{cury}){\line(4,3){\value{temp}}}
       \put(\value{curx},\value{cury}){\line(-1,3){\value{tapeboxhight}}}
       \put(\value{curx},\value{cury}){\line(1,3){\value{tapeboxhight}}}
       \setcounter{cury}{4*\value{tapeboxhight}}
       \setcounter{temp}{4*\value{tapeboxhight}}
       \put(\value{curx},\value{cury}){\line(-4,-3){\value{temp}}}
       \put(\value{curx},\value{cury}){\line(4,-3){\value{temp}}}
       \put(\value{curx},\value{cury}){\line(-1,-3){\value{tapeboxhight}}}
       \put(\value{curx},\value{cury}){\line(1,-3){\value{tapeboxhight}}}
    \end{picture}%
   }

\newcommand{\katajakoskenniemi}%
   {
    \setcounter{picwidth}{10*\value{tapeboxhight}}%
    \setcounter{pichight}{9*\value{tapeboxhight}}%
    \setcounter{lift}{\value{pichight}/-2}%
    \rule[\the\value{lift}pt]{0 pt}{\the\value{pichight}pt}%
    \begin{picture}(\the\value{picwidth},0)(0,-\the\value{lift})%
       \put(0,0){\makebox(\value{picwidth},\value{tapeboxhight})
             {Surface Representation}}%
       \setcounter{cury}{4*\value{tapeboxhight}}
       \setcounter{temp}{8*\value{tapeboxhight}}
       \put(\value{tapeboxhight},\value{cury})
             {\makebox(\value{temp},\value{tapeboxhight})
             {Lexical Representation}}%
       \setcounter{cury}{8*\value{tapeboxhight}}
       \put(0,\value{cury}){\makebox(\value{picwidth},\value{tapeboxhight})
             {Lexical Entries (Morphemes)}}%
       \setcounter{curx}{\value{picwidth}/2}
       \setcounter{cury}{\value{tapeboxhight}}
       \setcounter{delta}{2*\value{tapeboxhight}}
       \multiput(\value{curx},\value{cury})(0,\value{delta}){4}
          {\line(0,1){\value{tapeboxhight}}}
       \setcounter{cury}{2*\value{tapeboxhight}+\value{tapeboxhight}/2}
       \setcounter{delta}{4*\value{tapeboxhight}}
       \setcounter{fstwidth}{10*\value{tapeboxhight}}
       \multiput(\value{curx},\value{cury})(0,\value{delta}){2}
          {\oval(\value{fstwidth},\value{tapeboxhight})}
       \put(\value{curx},\value{cury}){\makebox(0,0){\sc Two-Level Rules}}%
       \setcounter{cury}{6*\value{tapeboxhight}+\value{tapeboxhight}/2}
       \put(\value{curx},\value{cury}){\makebox(0,0){\sc Lexicon Component}}%
    \end{picture}%
   }

\newcommand{\environbar}{\underline{\hspace*{1.5em}}\ }

\newcommand{\phonrule}[4]%
   {#1 {}$\rightarrow${} #2 / #3 \environbar #4}

\newcommand{\tlr}[4]%
   {#1 \ \ \ #2 \ \ \ #3%
    \ifthenelse{\equal{#4}{}}{}%
               {\hspace{0.25in}{\sf where} #4}%
   }

\newcommand{\tlrt}[8]%
   {\begin{tabular}{cccccc}
      {}#5&-&#6&-&#7&#4 \\
      {}#1&-&#2&-&#3& \\ 
      \end{tabular}%
    \ifthenelse{\equal{#8}{}}{}%
               {\\ {}\hspace{1cm} {\sf where} #8}%
   \vspace{.1in}}


\newcommand{\A}{\ejective}  
\newcommand{\h}{\crossh}    
\newcommand{\CC}{\reveject} 
\newcommand{\sh}{\v{s}}	    
\newcommand{\J}{\^{\j}}     
\newcommand{\TT}{\d{t}}
\newcommand{\Ss}{\d{s}}

\newcommand{\br}{\b{b}}   
\newcommand{\gr}{\b{g}}
\newcommand{\dr}{\b{d}}
\newcommand{\kr}{\b{k}}
\newcommand{\pr}{\b{p}}
\newcommand{\tr}{\b{t}}
\newcommand{\e}{\schwa}

\newcommand{\sem}[1]%
   {
    /{\em \transcribe#1||END}/}

\newcommand{\sema}[1]%
   {
    {\em \transcribe#1||END}}

\def\transcribe#1#2#3END
   {
    \ifthenelse{\equal{#2}{.}}%
               {\b{#1}%
                \transcribe#3END}%
               {
                \ifthenelse{\equal{#1}{A}}{\A}{}%
                \ifthenelse{\equal{#1}{b}}{b}{}%
    		\ifthenelse{\equal{#1}{g}}{g}{}%
    		\ifthenelse{\equal{#1}{d}}{d}{}%
    		\ifthenelse{\equal{#1}{h}}{h}{}%
    		\ifthenelse{\equal{#1}{w}}{w}{}%
    		\ifthenelse{\equal{#1}{z}}{z}{}%
    		\ifthenelse{\equal{#1}{H}}{\d{h}}{}%
    		\ifthenelse{\equal{#1}{T}}{\d{t}}{}%
    		\ifthenelse{\equal{#1}{y}}{y}{}%
    		\ifthenelse{\equal{#1}{k}}{k}{}%
    		\ifthenelse{\equal{#1}{l}}{l}{}%
    		\ifthenelse{\equal{#1}{m}}{m}{}%
    		\ifthenelse{\equal{#1}{n}}{n}{}%
    		\ifthenelse{\equal{#1}{s}}{s}{}%
    		\ifthenelse{\equal{#1}{c}}{\reveject}{}%
    		\ifthenelse{\equal{#1}{p}}{p}{}%
    		\ifthenelse{\equal{#1}{S}}{\d{s}}{}%
    		\ifthenelse{\equal{#1}{q}}{q}{}%
    		\ifthenelse{\equal{#1}{r}}{r}{}%
    		\ifthenelse{\equal{#1}{W}}{\v{s}}{}%
    		\ifthenelse{\equal{#1}{t}}{t}{}%
    		\ifthenelse{\equal{#1}{a}}{a}{}%
    		\ifthenelse{\equal{#1}{O}}{\={a}}{}%
    		\ifthenelse{\equal{#1}{e}}{e}{}%
    		\ifthenelse{\equal{#1}{E}}{\={e}}{}%
    		\ifthenelse{\equal{#1}{i}}{i}{}%
     		\ifthenelse{\equal{#1}{I}}{\={\i}}{}%
    		\ifthenelse{\equal{#1}{o}}{o}{}%
    		\ifthenelse{\equal{#1}{u}}{u}{}%
    		\ifthenelse{\equal{#1}{U}}{\={u}}{}%
    		\ifthenelse{\equal{#1}{j}}{\^{\j}}{}%
    		\ifthenelse{\equal{#1}{v}}{\raisebox{.75ex}{{\small{\em e}}}}{}%
    		\ifthenelse{\equal{#1}{-}}{\ }{}%
    		\ifthenelse{\equal{#2}{|}}{}{\transcribe#2#3END}}}

%

\newcounter{examplectr}
\newcounter{subexamplectr}
%
\newenvironment{ex}%
   {\addtocounter{examplectr}{1}%
     \setcounter{subexamplectr}{0}%
\vspace{.1in}     \begin{list}%
       {(\arabic{examplectr})}%
       {\setlength{\topsep}{0in}%
	\setlength{\leftmargin}{0.45in}%
	\setlength{\labelsep}{0.075in}}%
       \item \begin{minipage}[t]{5.5in}%
   }%
   {\end{minipage}%
    \end{list}\vspace{.1in}}%
%
\newenvironment{subex}%
   { \addtocounter{subexamplectr}{1}
     \begin{list}
       {\alph{subexamplectr}.}%
       {\setlength{\topsep}{-\parskip}
	\setlength{\leftmargin}{0.175in}
	\setlength{\labelsep}{0.075in}}
       \item
   }%
   {\end{list}}
%
\newcommand{\exnum}[2]{\addtocounter{examplectr}{#1}(\arabic{examplectr}{#2})\addtocounter{examplectr}{-#1}}


\author{Tanya Bowden \thanks{\hspace{.1in}Supported by a British Telecom
Scholarship,
administered by the Cambridge Commonwealth Trust in conjunction with the
Foreign and
Commonwealth Office.}
\and George Anton Kiraz \thanks{\hspace{.1in}Supported by a Benefactor
Studentship
from St John's College.}\\
 University of Cambridge \\
 Computer Laboratory \\
 Pembroke Street, Cambridge CB2 3QG \\
 {\tt \{Tanya.Bowden, George.Kiraz\}@cl.cam.ac.uk} \\
 {\tt http://www.cl.cam.ac.uk/users/\{tgb1000, gk105\}} \\
  \mbox{}}

\title{A Morphographemic Model for Error Correction in Nonconcatenative
Strings}
\maketitle

\begin{abstract}
This paper introduces a spelling correction system which integrates
seamlessly with morphological analysis using a multi-tape formalism.
Handling of various Semitic error problems is illustrated, with
reference to Arabic and Syriac examples.  The model handles errors
vocalisation, diacritics, phonetic syncopation and morphographemic
idiosyncrasies, in addition to Damerau errors.  A complementary
correction strategy for morphologically sound but morphosyntactically
ill-formed words is outlined.
\end{abstract}


\section{Introduction}
\label{intro}

Semitic is known amongst computational linguists, in particular
computational morphologists, for its highly inflexional morphology.
Its root-and-pattern phenomenon not only poses difficulties for a
morphological system, but also makes error detection a difficult
task. This paper aims at presenting a morphographemic model which can
cope with both issues.

The following convention has been adopted. Morphemes are represented
in braces, \{ \}, surface (phonological) forms in solidi, / /, and
orthographic strings in acute brackets, \lab\ \rab. In examples of
grammars, variables begin with a capital letter. Cs denote consonants,
Vs denote vowels and a bar denotes complement. An asterisk, *,
indicates ill-formed strings.


The difficulties in morphological analysis and error detection in
Semitic arise from the following facts:
\begin{itemize}
   \item {\bf Non-Linearity} A Semitic stem consists of a {\bf root}
         and a {\bf vowel melody}, arranged according to a {\bf
         canonical pattern}. For example, Arabic /kuttib/ `caused to
         write - perfect passive' is composed from the root morpheme
         \{ktb\} `notion of writing' and the vowel melody morpheme
         \{ui\} `perfect passive'; the two are arranged according to
         the pattern morpheme \{CVCCVC\} `causative'. This phenomenon
         is analysed by \cite{McCarthy:81} along the lines of
         autosegmental phonology \cite{Goldsmith:76}. The analysis
         appears in \exnum{+1}{}.\footnote{We have used the CV model
         to describe pattern morphemes instead of prosodic terms
         because of its familiarity in the computational linguistics
         literature. For the use of moraic and affixational models in
         handling Arabic morphology computationally, see
         \cite{Kiraz:forth}.}  \begin{ex} {\sc Derivation of} /kuttib/
         \\ /kuttib/ =
         \autoseg{k-t--b|CVCCVC|-u--i-}{21x,23x,21x,21l,23x,21x}
         \end{ex}

   \item {\bf Vocalisation} Orthographically, Semitic texts appear in
         three forms: (i) {\bf consonantal texts} do not incorporate
         any short vowels but {\em matres
         lectionis},\footnote{`Mothers of reading', these are
         consonantal letters which play the role of long vowels, and
         are represented in the pattern morpheme by VV (e.g. /aa/,
         /uu/, /ii/).  {\em Matres lectionis} cannot be omitted from
         the orthographic string.} e.g. Arabic \lab ktb\rab\ for /katab/,
         /kutib/ and /kutub/, but \lab kaatb\rab\ for /kaatab/ and
         /kaatib/; (ii) {\bf partially vocalised texts} incorporate
         some short vowels to clarify ambiguity, e.g. \lab kutb\rab\
         for /kutib/ to distinguish it from /katab/; and (iii) {\bf
         vocalised texts} incorporate full vocalisation, e.g. \lab
         tada\h ra\J\rab\ for /tada\h ra\J/.

   \item {\bf Vowel and Diacritic Shifts} Semitic languages employ a
         large number of diacritics to represent {\em enter alia}
         short vowels, doubled letters, and nunation.\footnote{When
         indefinite, nouns and adjectives end in a {\em phonetic} [n]
         which is represented in the {\em orthographic} form by
         special diacritics.} Most editors allow the user to enter
         such diacritics above and below letters. To speed data entry,
         the user usually enters the base characters (say a paragraph) and
         then goes back and enters the diacritics. A common mistake is
         to place the cursor one extra position to the left when
         entering diacritics. This results in the vowels being shifted
         one position, e.g. *\lab wkatubi\rab\ instead of \lab
         wakutib\rab.

   \item {\bf Vocalisms} The quality of the perfect and imperfect
         vowels of the basic forms of the Semitic verbs are
         idiosyncratic. For example, the Syriac root \{ktb\} takes
         the perfect vowel {\em a}, e.g. /ktab/, while the root
         \{n\h t\} takes the vowel {\em e}, e.g. /n\h et/. It is
         common among learners to make mistakes such as */kteb/ or
         */n\h at/.

   \item {\bf Phonetic Syncopation} A consonantal segment may be
         omitted from the {\em phonetic} surface form, but maintained
         in the {\em orthographic} surface from. For example, Syriac
         \lab md\={\i}nt\^{a}\rab `city' is pronounced /md\={\i}t\^{a}/.

   \item {\bf Idiosyncrasies} The application of a morphographemic
          rule may have constraints as on which lexical morphemes it
          may or may not apply. For example, the glottal stop [\A]
          at the end of a stem may become [w] when followed by the
          relative adjective morpheme \{iyy\}, as in Arabic
          /samaa\A+iyy/ $\rightarrow$ /samaawiyy/ `heavenly', but
          /hawaa\A+iyy/ $\rightarrow$ /hawaa\A iyy/ `of
          air'. \\

    \item {\bf Morphosyntactic Issues} In broken plurals, diminutives
          and deverbal nouns, the user may enter a morphologically
          sound, but morphosyntactically ill-formed word. We shall
          discuss this in more detail in
          section~\ref{bp}.\footnote{For other issues with respect to
          syntactic dependencies, see \cite{Daud:90}.}
\end{itemize}
To the above, one adds language-independent issues in spell checking
such as the four Damerau transformations: omission, insertion,
transposition and substitution \cite{Damerau:64}.

\section{A Morphographemic Model}
\label{model}

This section presents a morphographemic model which handles error
detection in non-linear strings. Subsection~\ref{formalism} presents
the formalism used, and subsection~\ref{model2} describes the model.

\subsection{The Formalism}
\label{formalism}

In order to handle the non-linear phenomenon of Arabic, our model
adopts the two-level formalism presented by \cite{Pulman:93}, with the
multi tape extensions in \cite{Kiraz:94Coling}. Their formalism appears
in \exnum{+1}{}.
\begin{ex}
   {\sc Two-Level Formalism} \vspace{.1in} \\
      \tlrt{LSC}{\sc Surf}{RSC}{\{$\Rightarrow,\Leftrightarrow$\}}
                {LLC}{\sc Lex}{RLC}{} \\
where
\\
\begin{tabular}{lll}
   {\sc LLC} &=& left lexical context\\
   {\sc Lex} &=& lexical form \\
   {\sc RLC} &=& right lexical context \\
   {\sc LSC}  &=& left surface context  \\
   {\sc Surf} &=& surface form          \\
   {\sc RSC}  &=& right surface context
   \end{tabular}
\end{ex}
The special symbol * is a wildcard matching any context, with no
length restrictions.  The operator $\Leftrightarrow$ caters for obligatory
rules. A lexical string maps to a surface string iff they can be
partitioned into pairs of lexical-surface subsequences, where each
pair is licenced by a $\Rightarrow$ or $\Leftrightarrow$ rule, and no
partition violates a $\Leftrightarrow$ rule. In the multi-tape
version, lexical expressions (i.e. {\sc LLC}, {\sc Lex} and {\sc RLC})
are {\em n}-tuple of regular expressions of the form (x$_1$, x$_2$,
$\ldots$, x$_n$): the {\em i}th expression refers to symbols on the
{\em i}th tape; a nill slot is indicated by \estr.\footnote{Our
implementation interprets rules directly; hence, we allow \estr. If
the rules were to be compiled into automata, a genuine symbol, e.g.~0,
must be used. For the compilation of our formalism into automata, see
\cite{Kiraz-Evans:forth}.} Another extension is giving {\sc LLC} the
ability to contain ellipsis, $\ldots$ , which indicates the (optional)
omission from {\sc LLC} of tuples, provided that the tuples to the
left of $\ldots$ are the first to appear on the left of {\sc Lex}.

In our morphographemic model, we add a similar formalism for
expressing error rules \exnum{+1}{}.
\begin{ex}
   {\sc Error Formalism} \\
   ErrSurf $\Rightarrow$ Surf  \\
   {\sc \{ PLC - PRC \} }
where
\\
\begin{tabular}{lll}
   {\sc PLC} &=& partition left context \\
             & & (has been done)\\
   {\sc PRC} &=& partition right context \\
             & & (yet to be done)\\
   \end{tabular}
\end{ex}

The error rules capture the correspondence between the error surface
and the correct surface, given the surrounding partition into surface
and lexical contexts.  They happily utilise the multi-tape format and
integrate seamlessly into morphological analysis.  PLC and PRC above
are the left and right contexts of both the lexical and (correct)
surface levels.  Only the $\Rightarrow$ is used (error is not obligatory).

\subsection{The Model}
\label{model2}

\subsubsection{Finding the error}
Morphological analysis is first called with the assumption that
the word is free of errors.  If this fails, analysis is attempted
again without the `no error' restriction.
The error rules are then considered when ordinary
morphological rules fail.  If no error rules succeed, or lead to
a successful partition of the word, analysis backtracks to try the
error rules at successively earlier points in the word.

For purposes of simplicity and because on the whole is it likely that
words will contain no more than one error \cite{Damerau:64,Pollock:83}, normal
`no error' analysis
usually resumes if an error rule succeeds.  The exception occurs with
a vowel shift error (\S\ref{shift}).  If this error rule succeeds, an
expectation of further shifted vowels is set up, but no other error
rule is allowed in the subsequent partitions.  For this reason rules
are marked as to whether they can occur more than once.

\subsubsection{Suggesting a correction}
Once an error rule is selected, the corrected surface is substituted for
the error surface, and normal analysis continues - at the same position.
The substituted surface may be in the form of a variable, which
is then ground by the normal analysis sequence of lexical matching over
the lexicon tree. In this way only lexical words are considered, as
the variable letter can only be instantiated to letters branching out
from the current position on the lexicon tree.  Normal prolog backtracking
to explore alternative rules/lexical branches applies throughout.

\section{Error Checking in Arabic}
\label{arabic}

We demonstrate our model on the Arabic verbal stems shown in \exnum{+1}{}
\cite{McCarthy:81}.
Verbs are classified according to their {\bf measure} (M): there are 15
trilateral measures and 4 quadrilateral ones.  Moving horizontally
across the table, one notices a change in vowel melody (active \{a\},
passive \{ui\}); everything else remains invariant.  Moving
vertically, a change in canonical pattern occurs; everything else
remains invariant.

Subsection~\ref{tlr} presents a simple two-level grammar which
describes the above data. Subsection~\ref{erules} presents error
checking.

\begin{ex}
   {\sc Arabic Verbal Stems} \\
   \begin{tabular}{rll}
     Measure & Active   & Passive\\
   \hline
   1  & katab    & kutib    \\
   2  & kattab   & kuttib   \\
   3  & kaatab   & kuutib    \\
   4  & \A aktab & \A uktib  \\
   5  & takattab & tukuttib \\
   6  & takaatab & tukuutib \\
   7  & nkatab   & nkutib    \\
   8  & ktatab   & ktutib   \\
   9  & ktabab   &          \\
   10 & staktab  & stuktib  \\
   11 & ktaabab  \\
   12 & ktawtab  \\
   13 & ktawwab  \\
   14 & ktanbab  \\
   15 & ktanbay  \\
   Q1 & da\h ra\J & du\h ri\J \\
   Q2 & tada\h ra\J & tudu\h ri\J\\
   Q3 & d\h anra\J & d\h unri\J \\
   Q4 & d\h ar\J a\J & d\h ur\J i\J
   \end{tabular}
\end{ex}

\subsection{Two-Level Rules}
\label{tlr}

The lexical level maintains three lexical tapes
\cite{Kay:87,Kiraz:94Coling}: pattern tape, root tape and vocalism
tape; each tape scans a lexical tree. Examples of pattern morphemes
are: \{c$_1$v$_1$c$_2$v$_1$c$_3$\} (M 1),
\{c$_1$c$_2$v$_1$nc$_3$v$_2$c$_4$\} (M Q3). The root morphemes
are \{ktb\} and \{d\h r\J\}, and the vocalism morphemes are \{a\}
(active) and \{ui\} (passive).

The following two-level grammar handles the above data. Each lexical
expression is a triple; lexical expressions with one symbol assume
\estr\ on the remaining positions.
\begin{ex}
   {\sc General Rules} \vspace{.1in}\\
         \vspace{.051in}R0: \tlrt{*}{X}{*}{$\Rightarrow$}{*}{X}{*}{} \\
         \vspace{.051in}R1: \tlrt{*}{C}{*}{$\Rightarrow$}
                                {*}{(P$_c$, C, \estr)}{*}
                                {} \\
         \vspace{.051in}R2: \tlrt{*}{V}{*}{$\Rightarrow$}
                                {*}{(P$_v$, \estr, V)}{*}
                                {} \\
   where P$_c$ $\in$ \{c$_1$, c$_2$, c$_3$, c$_4$\}, \\
         P$_v$ $\in$ \{v$_1$, v$_2$\},\\
\end{ex}

\exnum{0}{} gives three general rules: R0
allows any character on the first lexical tape to surface,
e.g. infixes, prefixes and suffixes.  R1 states that any P $\in$
\{c$_1$, c$_2$, c$_3$, c$_4$\} on the first (pattern) tape and C on
the second (root) tape with no transition on the third (vocalism) tape
corresponds to C on the surface tape; this rule sanctions consonants.
Similarly, R2 states that any P $\in$ \{v$_1$, v$_2$\} on the pattern
tape and V on vocalism tape with no transition on the root tape
corresponds to V on the surface tape; this rule sanctions vowels.
\begin{ex}
   {\sc Boundary Rules}  \vspace{.1in}\\
         \vspace{.051in}R3: \tlrt{*}{\estr}{*}{$\Rightarrow$}
                              {(B, \estr, \estr)}{+}{*}{} \\
         \vspace{.051in}R4: \tlrt{*}{\estr}{*}{$\Rightarrow$}
                              {(B,*,*)}{(+,+,+)}{*}{} \\
   where B $\neq$ +
\end{ex}
\exnum{0}{} gives two boundary rules: R3 is used for non-stem
morphemes, e.g.~prefixes and suffixes. R4 applies to stem morphemes
reading three boundary symbols simultaneously; this marks the end of a
stem. Notice that LLC ensures that the right boundary rule is invoked
at the right time.

Before embarking on the rest of the rules, an illustrated example
seems in order. The derivation of /d\h unri\J a/ (M Q5,
passive), from the three morphemes
\{c$_1$c$_2$v$_1$nc$_3$v$_2$c$_4$\}, \{d\h r\J\} and \{ui\}, and the
suffix \{a\} `3rd person' is illustrated in \exnum{+1}{}.
\begin{ex}
   {\sc Derivation of M Q3} + \{a\} \\
   \mttlm{d{\h}unri{\J}{}a-:surface tape}{1120121403}%
         {{c$_1$}{c$_2$}{v$_1$}n{c$_3$}{v$_2$}{c$_4$}+a+:pattern tape|%
          d{\h}--r-{\J}+--:root tape|%
          --u--i-+--:vocalism tape}
\end{ex}
The numbers between the surface tape and the lexical tapes indicate the rules
which sanction the moves.
\begin{ex}
   {\sc Spreading Rules} \vspace{.1in} \\
         \vspace{.051in}R5: \tlrt{*}{C}{*}{$\Rightarrow$}
                              {(P$_1$, C, \estr) $\cdots$}{P}{*}
                              {} \\
         \vspace{.051in}R6: \tlrt{*}{V}{*}{$\Rightarrow$}
                              {(v$_1$, \estr, V) $\cdots$}{v$_1$}{*}{} \\
   where  P$_1$ $\in$ \{c$_2$, c$_3$, c$_4$\}
\end{ex}
Resuming the description of the grammar, \exnum{0}{} presents spreading
rules. Notice the use of ellipsis to indicate that there can be tuples
separating {\sc Lex} and LLC, as far as the tuples in LLC are the nearest ones
to {\sc Lex}. R5 sanctions the spreading (and gemination) of consonants. R6
sanctions the spreading of the first vowel. Spreading examples appear in
\exnum{+1}{}.
\begin{ex}
   {\sc Derivation of M 1- M 3}
   \vspace{.1in}
   \begin{subex}
      /katab/ = \mttlm{katab{}:ST}{121614}%
            {{c$_1$}{v$_1$}{c$_2$}{v$_1$}{c$_3$}+:PT|k-t-b+:RT|-a---+:VT} \\
   \end{subex}
   \vspace{.1in}
   \begin{subex}
      /kattab/ = \mttlm{kattab{}:ST}{1215614}%
            {{c$_1$}{v$_1$}{c$_2$}{c$_2$}{v$_1$}{c$_3$}+:PT|%
            k-t--b+:RT|-a----+:VT} \\
   \end{subex}
   \vspace{.1in}
   \begin{subex}
      /kaatab/ = \mttlm{kaatab{}:ST}{1261614}%
            {{c$_1$}{v$_1$}{v$_1$}{c$_2$}{v$_1$}{c$_3$}+:PT|%
            k--t-b+:RT|-a----+:VT}
   \end{subex}
\end{ex}

The following rules allow for the different possible orthographic
vocalisations in Semitic texts: \vspace{.05in} \\
   R7~\tlrt{*}{\estr}{*}{$\Rightarrow$}
           {\small ($\overline{\mbox{V}}$, \estr,\estr)}
           {\small (V, \estr, \estr)}
           {\small($\overline{\mbox{V}}$, \estr, \estr)} {} \\
   R8~\tlrt{*}{\estr}{*}{$\Rightarrow$}
           {\small (P$_{c1}$, C1, \estr)}
           {\small (P, \estr, V)}
           {\small (P$_{c2}$, C2, \estr)} {} \\
   R9~\tlrt{*}{\estr}{*}{$\Rightarrow$}
           {\small $\lambda$}
           {\small(v$_1$,\estr,\estr)}
           {\small $\rho$} {} \\
where $\lambda$ = (v$_1$,\estr,V)$\cdots$(P$_{c1}$,C1,\estr) and
$\rho$ = (P$_{c2}$,C2,\estr).

R7 and R8 allow the optional deletion of short vowels in non-stem and
stem morphemes, respectively; note that the lexical contexts make sure
that long vowels are not deleted. R9 allows the optional deletion of a
short vowel what is the cause of spreading.
For example the rules sanction both /katab/ (M~1, active) and /kutib/
(M~1, passive) as interpretations of \lab ktb\rab\ as showin in
\exnum{+1}{}.

\subsection{Error Rules}
\label{erules}

Below are outlined error rules resulting from peculiarly Semitic
problems.  Error rules can also be constructed in a similar vein to
deal with typographical Damerau error (which also take care of the
issue of wrong vocalisms).

\begin{ex}
   {\sc Two-Level Derivation of M 1}
   \vspace{.1in}
   \begin{subex}
       /katab/ = \mttlm{k{}t{}b{}:ST}{181914}%
            {{c$_1$}{v$_1$}{c$_2$}{v$_1$}{c$_3$}+:PT|k-t-b+:RT|-a---+:VT}
   \end{subex}
   \vspace{.1in}
   \begin{subex}
      /kutib/ = \mttlm{k{}t{}b{}:ST}{181914}%
            {{c$_1$}{v$_1$}{c$_2$}{v$_1$}{c$_3$}+:PT|k-t-b+:RT|-u-i-+:VT}
   \end{subex}
\end{ex}

\subsubsection{Vowel Shift}
\label{shift}
A vowel shift error rule will be tried with a partition on a (short)
vowel which is not an expected (lexical) vowel at that position.
Short vowels can legitimately be omitted from an orthographic representation
-~it is this fact which contributes to the problem of vowel shifts.
A vowel is considered shifted if the same vowel has been omitted
earlier in the word.  The rule deletes the vowel from the surface. Hence
in the next pass of (normal) analysis, the partition is analysed as
a legitimate omission of the {\em expected} vowel.  This prepares for the next
shifted vowel to be treated in exactly the same way as the first. The
expectation of this reapplication is allowed for in reap~= y.

\begin{ex}
  E0: \tlr{X}{$\Rightarrow$}{\estr}{reap = y} \\
 \{ [om\_stmv,\estr,(*,*,X)] $\cdots$ - * \}    \\
\\

  E1: \tlr{X}{$\Rightarrow$}{\estr}{reap = y} \\
 \{ [*,*,(v1,\estr,X)] $\cdots$ [om\_sprv,\estr,(*,*,\estr)] $\cdots$ - * \} \\
\end{ex}


In the rules above, `X' is the shifted vowel.  It is deleted from the surface.
The partition contextual tuples consist of {\sc [Rule Name, Surf, Lex]}.  The
{\sc Lex} element is a tuple itself of {\sc [Pattern, Root, Vocalism]}.
In E0 the shifted vowel was analysed earlier as an omitted stem vowel
(om\_stmv),
whereas in E1 it was analysed earlier as an omitted spread vowel (om\_sprv).
The surface/lexical restrictions in the contexts could be written out
in more detail, but both rules make use of the fact that those
contexts are analysed by other partitions, which check that they meet the
conditions for an omitted stem vowel or omitted spread vowel.

For example, *\lab d\h ru\J i\rab\ will be interpreted as \lab du\h ri\J\rab .
The `E0's on the rule number line indicate where the vowel shift rule was
applied to replace an error surface vowel with \estr.  The error surface vowels
are written in italics.

\begin{ex}
   {\sc Two-Level Analysis of} *\lab d\h ru\J i\rab\  \\
       \mttlm{d{}{\h}r{\it u}{}{\J}{\it i}{}:ST}{1811{E0}81{E0}4}%
{{c$_1$}{v$_1$}{c$_2$}{c$_3$}{}
{v$_2$}{c$_4$}{}+:PT|d-{\h}r--{\J}-+:RT|-u---i--+:VT}
\end{ex}


\subsubsection{Deleted Consonant}
Problems resulting from phonetic syncopation can be treated as
accidental omission of a consonant, e.g.~ *\lab md\={\i}t\^{a}\rab, \lab
md\={\i}nt\^{a}\rab.
\begin{ex}
  E2: \tlr{\estr}{$\Rightarrow$}{X}{cons(X),reap = n} \\
  \{ * - * \} \\
\end{ex}

\subsubsection{Deleted Long Vowel}
Although the error probably results from a different fault, a deleted
long vowel can be treated in the same way as a deleted consonant.
With current transcription practice, long vowels are commonly written
as two characters - they are possibly better represented as a single,
distinct character.

\begin{ex}
  E3: \tlr{\estr}{$\Rightarrow$}{XX}{vowel(X),reap = n} \\
  \{ * - * \} \\
\end{ex}

The form *\lab tuktib\rab\ can be interpreted as either \lab
tukuttib\rab\ with a deleted consonant (geminated `t') or \lab
tukuutib\rab\ with a deleted long vowel.

\begin{ex}
   {\sc Two-Level Analysis of} *\lab tuktib\rab
   \vspace{.1in}
    \begin{subex}
       M 5 =
       \mttlm{tuk{}t{}tib{}:ST}{02191{E2}1214}%
{t{v$_1$}{c$_1$}{v$_1$}
{c$_2$}{}{c$_2$}{v$_2$}{c$_3$}+:PT|--k-t---b+:RT|-u-----i-+:VT}
    \end{subex}
   \vspace{.1in}
    \begin{subex}
       M 6 = \mttlm{tuk{}uutib{}:ST}{021{E3}661214}%
{t{v$_1$}{c$_1$}{}{v$_1$}{v$_1$}{c$_2$}{v$_2$}
{c$_3$}+:PT|--k---t-b+:RT|-u-----i-+:VT}
    \end{subex}
\end{ex}


\subsubsection{Substituted Consonant}
One type of morphographemic error is that consonant substitution may
not take place before appending a suffix.  For example /samaa\A/
`heaven' + \{iyy\} `relative adjective' surfaces as \lab
samaawiyy\rab, where \A $\rightarrow$ w in the given context. A common
mistake is to write it as *\lab samma\A iyy\rab.

\begin{ex}
  E4: \tlr{\A}{$\Rightarrow$}{w}{reap = n} \\
  \{ * - {[glottal\_change, w,(P$_c$,\A,\estr)]} \}
\end{ex}

The `glottal\_change' rule would be a normal morphological spelling change
rule, incorporating contextual constraints (e.g. for the morpheme boundary)
as necessary.

 \section{Broken Plurals, Diminutive and Deverbal Nouns} \label{bp}

This section deals with morphosyntactic errors which are independent
of the two-level analysis. The data described below was obtained from
Daniel Ponsford (personal communication), based on \cite{Wehr:71}.

Recall that a Semitic stems consists of a root morpheme and a
vocalism morpheme arranged according to a canonical pattern morpheme.
As each root does not occur in all vocalisms and patterns, each
lexical entry is associated with a feature structure which indicates
{\em inter alia} the possible patterns and vocalisms for a particular
root. Consider the nominal data in \exnum{+1}{}.
\begin{ex}
   {\sc Broken Plurals} \\
   \begin{tabular}{ll}
   Singular & Plural Forms \\ \hline
   kadi\sh & kud\sh, *kidaa\sh \\
   kaafil & kuffal, *kufalaa\A, *kuffaal \\
   kafiil & kufalaa\A \\
   sahm & *\A ashaam, suhuum, \A ashum
   \end{tabular}
\end{ex}
Patterns marked with * are morphologically plausible, but do not occur
lexically with the cited nouns. A common mistake is to choose the wrong
pattern.

In such a case, the two-level model succeeds in finding two-level
analyses of the word in question, but fails when parsing the word
morphosyntactically: at this stage, the parser is passed a root,
vocalism and pattern whose feature structures do not unify.

Usually this feature-clash situation creates the problem of which
constituent to give preference to \cite{Langer:90}.  Here the vocalism
indicates the inflection (e.g.~broken plural) and the preferance of
vocalism pattern for that type of inflection belongs to the root.  For
example *\lab kidaa\sh \rab would be analysed as root \{kd\sh \} with
a broken plural vocalism.  The pattern type of the vocalism clashes
with the broken plural pattern that the root expects.  To correct, the
morphological analyser is executed in generation mode to generate the
broken plural form of \{kd\sh \} in the normal way.

The same procedure can be applied on diminutive and deverbal nouns.

 \section{Conclusion} \label{conclusion}

The model presented corrects errors resulting from combining
nonconcatenative strings as well as more standard morphological or
spelling errors.  It covers Semitic errors relating to vocalisation,
diacritics,
phonetic syncopation and morphographemic idiosyncrasies.  Morphosyntactic
issues of broken plurals, diminutives and deverbal nouns can be handled
by a complementary correction strategy which also depends on morphological
analysis.

Other than the economic factor, an important advantage of combining
morphological analysis and error detection/correction is the way the lexical
tree associated with the analysis can be used to determine correction
possibilities.
The morphological analysis proceeds by selecting rules that
hypothesise lexical strings for a given surface string. The rules are
accepted/rejected by checking that the lexical string(s) can extend
along the lexical tree(s) from the current position(s).  Variables
introduced by error rules into the surface string are then instantiated by
associating surface with lexical, and matching lexical strings to
the lexicon tree(s).  The system is unable to consider correction
characters that would be lexical impossibilities.


\section*{Acknowledgements}
The authors would like to thank their supervisor Dr~Stephen
Pulman. Thanks to Daniel Ponsford for providing data on the broken
plural and Nuha Adly Atteya for discussing Arabic examples.



\end{document}